%% file: Inan-Liu-Shehu-ArXiv-March01-2024.tex
\documentclass[letterpaper]{article} 
\usepackage[]{aaai24}  
\usepackage{times}  
\usepackage{helvet}  
\usepackage{courier}  
\usepackage[hyphens]{url}  
\usepackage{graphicx} 
\usepackage{natbib}  
\usepackage{caption} 
\usepackage{algorithm}
\usepackage{algorithmic}

\usepackage{subcaption}
\usepackage{booktabs} 

\usepackage{newfloat}
\usepackage{listings}
\DeclareCaptionStyle{ruled}{labelfont=normalfont,labelsep=colon,strut=off} 
\floatstyle{ruled}
\newfloat{listing}{tb}{lst}{}
\floatname{listing}{Listing}
\title{Beyond Single-Model Views for Deep Learning:\\ Optimization versus Generalizability of Stochastic Optimization Algorithms}

\author {
    Toki Tahmid Inan\textsuperscript{\rm 1},
    Mingrui Liu\textsuperscript{\rm 2},
    Amarda Shehu\textsuperscript{\rm 3}
}
\affiliations {
    Department of Computer Science, George Mason University,\\Virginia, USA\\
    \textsuperscript{\rm 1}tinan@gmu.edu, \textsuperscript{\rm 2}mingruil@gmu.edu, \textsuperscript{\rm 3}ashehu@gmu.edu
}

\usepackage{bibentry}

\usepackage{multirow}
\usepackage{amssymb,amsmath,amsthm,enumitem}

\input{math-symbols}

\usepackage[linewidth=1.2pt,linecolor=black]{mdframed}
\usepackage{algorithm}
\usepackage{algorithmic}

\begin{document}
\maketitle
\input{sections/Abstract}

\input{sections/Introduction}
\input{sections/Related-Work}

\input{sections/Benchmarking-Setup}

\input{sections/Basin-Hopping-Algorithms}

\input{sections/SyntheticLoss-Analysis}
\input{sections/Real-World-Analysis}

\input{sections/Limitations-Future-Work}
\input{sections/Conclusion}

\bibliography{references/Background-Refs,references/Shehu-Refs,references/Liu-Refs}

\end{document}

%% file: math-symbols.tex
\def \R {\mathbb{R}}

\def \w {\mathbf{w}}

\def \g {\mathbf{g}}

\def \g {\mathbf{g}}

\def \w {\mathbf{w}}

\def \R {\mathbb{R}}



%% file: sections/Abstract.tex
\begin{abstract}
Despite an extensive body of literature on deep learning optimization, our current understanding of what makes an optimization algorithm effective is fragmented. In particular, we do not understand well whether enhanced optimization translates to improved generalizability. Current research overlooks the inherent stochastic nature of stochastic gradient descent (SGD) and its variants, resulting in a lack of comprehensive benchmarking and insight into their statistical performance. This paper aims to address this gap by adopting a novel approach. Rather than solely evaluating the endpoint of individual optimization trajectories, we draw from an ensemble of trajectories to estimate the stationary distribution of stochastic optimizers. Our investigation encompasses a wide array of techniques, including SGD and its variants, flat-minima optimizers, and new algorithms we propose under the Basin Hopping framework. Through our evaluation, which encompasses synthetic functions with known minima and real-world problems in computer vision and natural language processing, we emphasize fair benchmarking under a statistical framework, comparing stationary distributions and establishing statistical significance. Our study uncovers several key findings regarding the relationship between training loss and hold-out accuracy, as well as the comparable performance of SGD, noise-enabled variants, and novel optimizers utilizing the BH framework. Notably, these algorithms demonstrate performance on par with flat-minima optimizers like SAM, albeit with half the gradient evaluations. We anticipate that our work will catalyze further exploration in deep learning optimization, encouraging a shift away from single-model approaches towards methodologies that acknowledge and leverage the stochastic nature of optimizers.
\end{abstract}

%% file: sections/Introduction.tex
\section{Introduction}
\label{sec:Intro}

While we now frame the training process during deep learning as the optimization of a typically complex, nonconvex objective/loss function, we still lack a comprehensive understanding and cannot guarantee the dynamics during training~\cite{PoggioLiao20}. We rely on gradient-descent algorithms originally developed and well-characterized for convex optimization. Certainly, stochastic gradient descent (SGD), a variant of the gradient descent (GD) algorithm for deep learning, has become the cornerstone optimization algorithm for training~\cite{BottouNocedal18}, and its empirically good performance has been reported in many papers across various application settings.

Growing theoretical work endeavors to understand when and why SGD and its variants work well or not. The focus often lies on the optimizers' ability to generalize effectively, matching their performance on testing data with that on training data~\cite{Chatterjee20}. The literature is extensive and sometimes presents contradictory findings. However, an increasingly popular avenue of research seeks to establish connections between flat, low-loss regions of the landscape and good generalization~\cite{KeskarNocedal16, BaldassiZecchina20, ForetNeyshabur21, BaldassiZecchina21, ZhaoHu22}, followed by the development of optimizers that bias their exploration of high-dimensional, nonconvex loss landscapes toward flat local minima~\cite{IzmailovGordon18, ForetNeyshabur21}. It's important to note that since all these algorithms are stochastic (whether through mini-batches or deliberate noise injection), no guarantee can be made regarding their ability to reach the global minimum.

The deep learning optimization literature reveals our fragmented understanding of what constitutes a good optimizer and, crucially, whether enhanced optimization performance translates to improved generalizability. While investigating the latter point is paramount, a common observation across the literature is the reporting of findings based on \emph{a single} model. Typically, this model represents either the point of convergence for the optimizer or the model with the lowest loss within a convergence window. However, this practice overlooks a fundamental characteristic inherent in SGD, its variants, and flat-minima optimizers: their intrinsic stochasticity. This paper bridges this gap by addressing the stochastic nature of deep learning optimizers. Specifically, the paper presents the following contributions:

\subsection{Contributions}
\begin{enumerate}

\item[1.] \textbf{From a single model to a model population perspective:} By viewing model training as an optimization journey guided by gradients and influenced by the stochastic nature of SGD and similar algorithms, we deepen our understanding of how optimization, generalization (as reflected by testing error), and the comparison of optimizers across a spectrum of models encountered during the loss landscape exploration, interrelate.

\item[2.] \textbf{Thorough evaluation across artificial and real-world scenarios:} Our approach includes a detailed comparison of optimizer performance on both artificially designed functions with known characteristics and real-world tasks in fields like computer vision and natural language processing. We introduce novel methods for comparing model populations and employ statistical significance tests to solidify our findings.

\item[3.] \textbf{Introduction of novel stochastic optimization algorithms within the Basin Hopping (BH) framework:} We explore a wide array of algorithms, including not just SGD and its noise-inclusive variants but also new gradient descent-based algorithms developed through the BH framework. This framework has inspired numerous novel stochastic optimization techniques across different fields. To facilitate further research, we are sharing the code for all algorithms, along with their hyperparameters and comparison methodologies.

\item[4.] \textbf{Establishing new benchmarks for stochastic optimizers:} We go beyond rate of convergence to generalization performance in a model population view. Taking into account the stochastic nature of optimizers, we introduce new benchmarking practices to better understand the relationship between better optimizers and better generalizability, as well as properly characterize the advantages of novel optimizers in the presence of complex, nonconvex loss functions.  

\end{enumerate}

%% file: sections/Related-Work.tex
\section{Background and Related Work}
\label{sec:RelatedWork}

\subsection{Stochastic Gradient Descent} 

Consider a multi-dimensional variable/parameter space $\w \in \R^p$ and a loss function $f(\w)$ that lifts the variable space into a landscape. At every iteration $t \in [T]$, where $T$ is a fixed budget, GD takes a discrete step in the direction of steepest descent and updates:
\vspace*{-1mm}
$$\w_t = \w_{t-1} - \eta \cdot \hat{\textbf{g}}$$ 
where $\hat{\textbf{g}}$ is the normalized gradient vector $\textbf{g} = \nabla{f(\w_{t-1})}$ of $f$ evaluated at $\w_{t-1}$; $\w_0$ (initial conditions) are sampled at random over the variable space. 

The "stochastic" in SGD is a key difference from GD and refers to the stochasticity of minibatches~\cite{LeNg11, DuchiSinger11, Zeiler12, KingmaBa14}; SGD minimizes the empirical loss: 
\vspace*{-1mm}
$$\frac{1}{|\mathcal{B}_t|} \sum_{i \in \mathcal{B}_t} f_i(\w_t)$$ 
where $f_i$ is the loss for a data point $i$ in the minibatch $\mathcal{B}_t$ drawn from the training dataset at iteration $t$ of training. The minibatch construct avoids storing all data in memory and extends SGD to online settings~\cite{BottouNocedal18, Shai12, LeNg11}. 

\subsection{Exploration versus Exploitation} 

The step size $\eta$ (in GD-based optimizers, including SGD) determines how much to ``walk" in the direction of the (negative) gradient; a large value risks overshooting and increasingly deviating away from $f$; a small value, while tracking $f$ more faithfully, risks premature convergence to a nearby minimum, possibly missing better ones. Work in~\cite{BaydinWood18} proposes optimizing $\eta$ via GD, and recent work automatically extends it to SGD variants~\cite{ChandraMeijer22}. 
However, for non-convex optimization, the ruggedness/multi-modality of the loss landscape (for which we have increasing evidence~\cite{LiGoldstein18, BosmanHelbig20, LiuBelkin20}) challenges controlling the balance between exploration (of the entirety of the landscape) and exploitation (of minima) through $\eta$ alone. The initial conditions $\w_0$ can also unduly impact the exploration.

\subsection{Noise-enabled Variants of SGD} 

Due to GD convergence to stationary points other than local minima (such as saddle points), early works proposed to incorporate randomness in the process, by injecting noise in the gradient~\cite{GeYuan15} or the model~\cite{JinJordan17}. Consider a noise vector $\eta$ drawn at random from $B_0(\rho)$ (a ball centered at the origin with radius $\rho$). In~\cite{GeYuan15}, this noise is added to the gradient prior to updating the model parameters, as shown in Algorithm~\ref{alg:NoiseInGradient}. Work in~\cite{JinJordan17} instead injects noise to the model parameters $\w$ directly, as shown in Algorithm~\ref{alg:NoiseInModel}, and conditionally, only after a certain number of iterations $\tau$ have been reached AND the magnitude of the gradient has become small. The first condition ensures that time is provided for exploitation via GD. The second condition identifies when a stationary point is reached. We rename these algorithms as NoiseInGradient-GD and NoiseInModel-GD and abbreviate them in the interest of space as NiG-GD and NiG-SGD. Note that the presentation here is for GD, but the SGD variants operate over the minibatches.

\begin{figure}[htbp]
\vspace*{-2mm}
\begin{mdframed}
\vspace*{-2mm}
\captionof{algorithm}{NiG-GD~\cite{GeYuan15}}
 \label{alg:NoiseInGradient}
 \begin{algorithmic}[1]
  \STATE {\bf Input:} $f(\w),T >0, \w, \eta, \rho$  
  \STATE {\bf Output:} $\w$
  \WHILE{$t \leq T$}
    \STATE $\g \leftarrow \nabla f(\w)$
    \STATE ${\bf \zeta} \in B_0(\rho)$ \hfill{$\triangleright${sample noise}}
    \STATE $\g \leftarrow \g + {\bf \zeta}$ \hfill{$\triangleright${add to gradient}}
    \STATE  $\w \gets \w - \eta \cdot \g$
  \ENDWHILE
 \end{algorithmic}
\end{mdframed}
\begin{mdframed}
\vspace*{-2mm}
\captionof{algorithm}{NiM-GD~\cite{JinJordan17}}
 \label{alg:NoiseInModel}
  \begin{algorithmic}[1]
  \STATE {\bf Input:} $f(\w), T >0, \epsilon \cong 0, \tau >0 ,\rho $ 
  \STATE {\bf Output:} $\w$ 
  \WHILE{$t \leq T$}
  \STATE ${\bf g} \leftarrow \nabla f(\w_t)$
   \IF{$\|{\bf g}\| < \epsilon $ and $ t > \tau$}
      \STATE ${\bf \zeta} \in B_0(\rho)$ \hfill{$\triangleright${sample noise}}
      \STATE $\w_t \gets \w_t+ \zeta$ \hfill{$\triangleright${to model}}
      \STATE ${\bf g} \leftarrow \nabla f(\w_t)$
   \ENDIF
   \STATE $\w \gets \w - \eta \cdot {\bf g}$
  \ENDWHILE
  \end{algorithmic}
\vspace*{-1mm}
\end{mdframed}
\vspace*{-4mm}
\end{figure}

Many follow-up works pursue variations of injecting noise. Work in~\cite{ZhouZhao19}, though limited to a simple two-layer convolutional neural network (CNN) model, shows that adding annealing noise to the gradient allows SGD to provably converge to a global optimum in polynomial time with arbitrary initialization. Work in~\cite{OrvietoBach22} connects injecting noise within GD with smoothing and regularization and shows that independent layer-wise perturbations circumvent the exploding variance term in over-parameterized models, yielding explicit regularization and better generalization. The stated motivation of noise-enabled optimizers is to escape saddle points.  There is a rich history and literature on noisy gradient methods based on the Langevin dynamics (LD)~\cite{Kennedy90,Neill11,Welling11,chaudhari2017entropysgd,MJordan18,chourasia2021differential}. Recent work~\cite{banerjee2022stability} additionally relaxes the Gaussian noise assumption within the LD framework. In this paper, we focus on the simplest noise-enabled variants of SGD, hoping to extend to LD-based ones in future work. For noise-enabled optimizers, we posit that it is useful to think of them as attempts to increase the exploration capability in a framework of exploration versus exploitation (as is common in stochastic optimization). While following a gradient increases exploitation, adding a perturbation to this via injecting noise in the gradient or directly the model enhances exploration.

The BH framework we re-introduce in this paper for deep learning optimizers additionally allows one to incorporate noise in a principled manner. For clarity, in this paper we limit our algorithmic exposition to the classic GD framework, but our evaluation setting considers the minibatch version of these algorithms (as SGD over GD).

\subsection{Flat-minima Optimizers for Deep Learning} 

Research on the benefit of seeking flat minima (with flatness loosely referring to the curvature of the neighborhood around a local minimum) is contradictory. One could summarize the current understanding as follows: Poorly generalizable local minima are sharp~\cite{KeskarNocedal16}.
SGD has an inherent bias to converge to flat local minima~\cite{SmithLe18}, and generalization can improve with further bias towards flat minima~\cite{IzmailovGordon18, ForetNeyshabur21}. However, sharp minima can generalize for deep nets~\cite{DinhBengio17} on a variety of tasks~\cite{KaddourKusner22}. 

Despite these contradictory findings, researchers are highly motivated to introduce novel optimization algorithms biased in some manner towards flat local minima. We single out here as a recent representative Sharpness Aware Minimization (SAM)~\cite{ForetNeyshabur21}. SAM minimizes the maximum loss around a neighborhood of the current SGD iterate but requires an additional forward/backward pass for each parameter update. As shown in Algorithm~\ref{alg:SAM}, rather than sampling a noise vector in $B_0(\rho)$, a deterministic vector $\zeta$ (of magnitude $\rho$) in the direction of the gradient is added to the model parameters at every iteration; There is no true noise injection, as $\rho$ is a user/input parameter. The gradient is also calculated twice (lines 4 and 8). SAM occupies its own category, given that it does not inject any noise but through a deterministic vector aims to get out of a stationary point.

\begin{figure}[htbp]
\begin{mdframed}
\vspace*{-2mm}
\captionof{algorithm}{SAM~\cite{ForetNeyshabur21}}
\label{alg:SAM}
\begin{algorithmic}[1]
\STATE {\bf Input:} $f(\w),T >0, \w, \eta, \rho > 0$
\STATE {\bf Output:} $\w$ 
\WHILE{$t \leq T$}
  \STATE ${\bf g}\gets \nabla f(\w)$
  \STATE ${\bf \hat{g}} \gets \frac{\bf g}{\|\bf g\|}$ \hfill{$\triangleright${normalize gradient}}
  \STATE ${\bf \zeta} \gets \rho \cdot {\bf \hat{g}}$ \hfill{$\triangleright${get perturb vector}}
  \STATE $\w\gets \w + {\bf \zeta}$ \hfill{$\triangleright${modify model}}
  \STATE ${\bf g}\gets \nabla f(\w)$ \hfill{$\triangleright${update gradient}}
  \STATE $ \w \gets \w - \eta \cdot {\bf g}$
\ENDWHILE
\end{algorithmic}
\end{mdframed}
\end{figure}

Many attempts have been made to understand SAM (as well as SGD and noise-enabled variants). Work in~\cite{BartlettOlivier22} provides bounds on SAM's rate of convergence and shows that, when applied with a convex quadratic objective, for most random initializations, SAM converges to a cycle that oscillates between either side of the minimum in the direction with the largest curvature. Comparison of SAM to another flat-minima optimizer, Stochastic Weight Averaging (SWA)~\cite{IzmailovGordon18} on diverse tasks (vision, NLP, etc.) shows no clear winner on convergence to flat minima, that SAM can converge to non-flat minima, and that non-flat minima sometimes have better generalization~\cite{KaddourKusner22}. 

%% file: sections/Benchmarking-Setup.tex
\section{Benchmarking Setup}
\label{BenchmarkingSetup}

\begin{figure*}[htbp]
\centering
\begin{tabular}{ccc}
\includegraphics[width=0.33\textwidth]{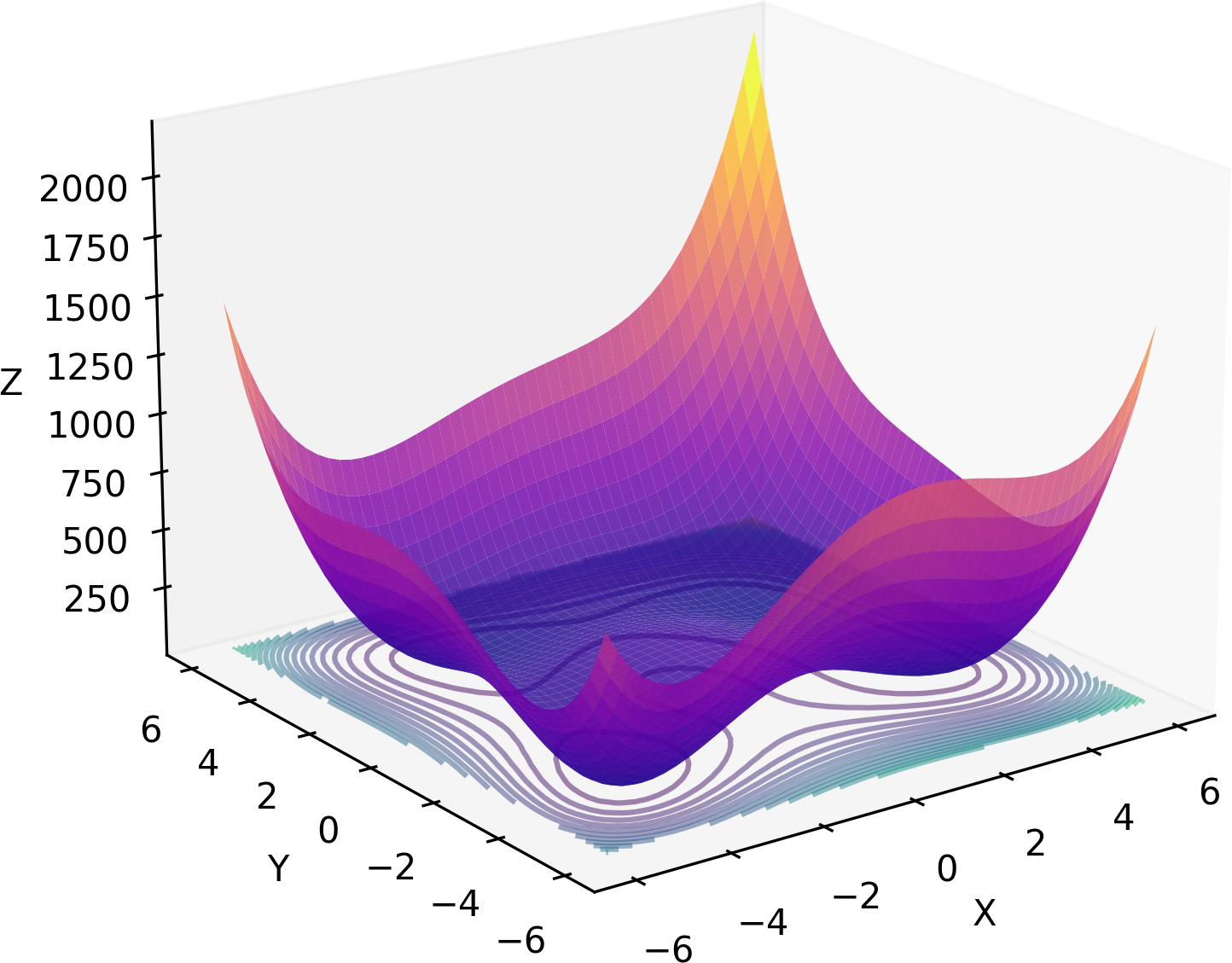} & 
\includegraphics[width=0.33\textwidth]{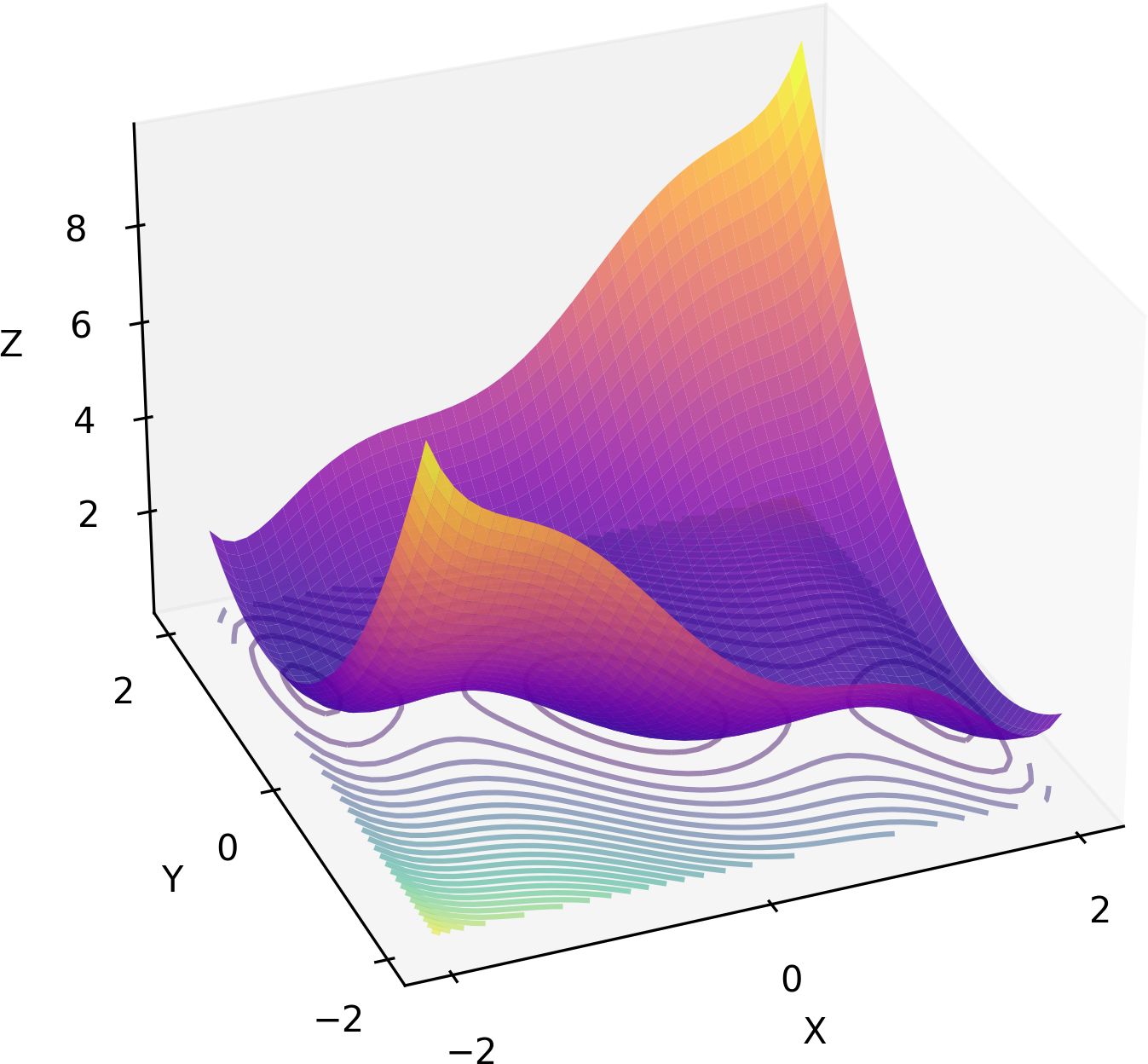} & 
\includegraphics[width=0.33\textwidth]{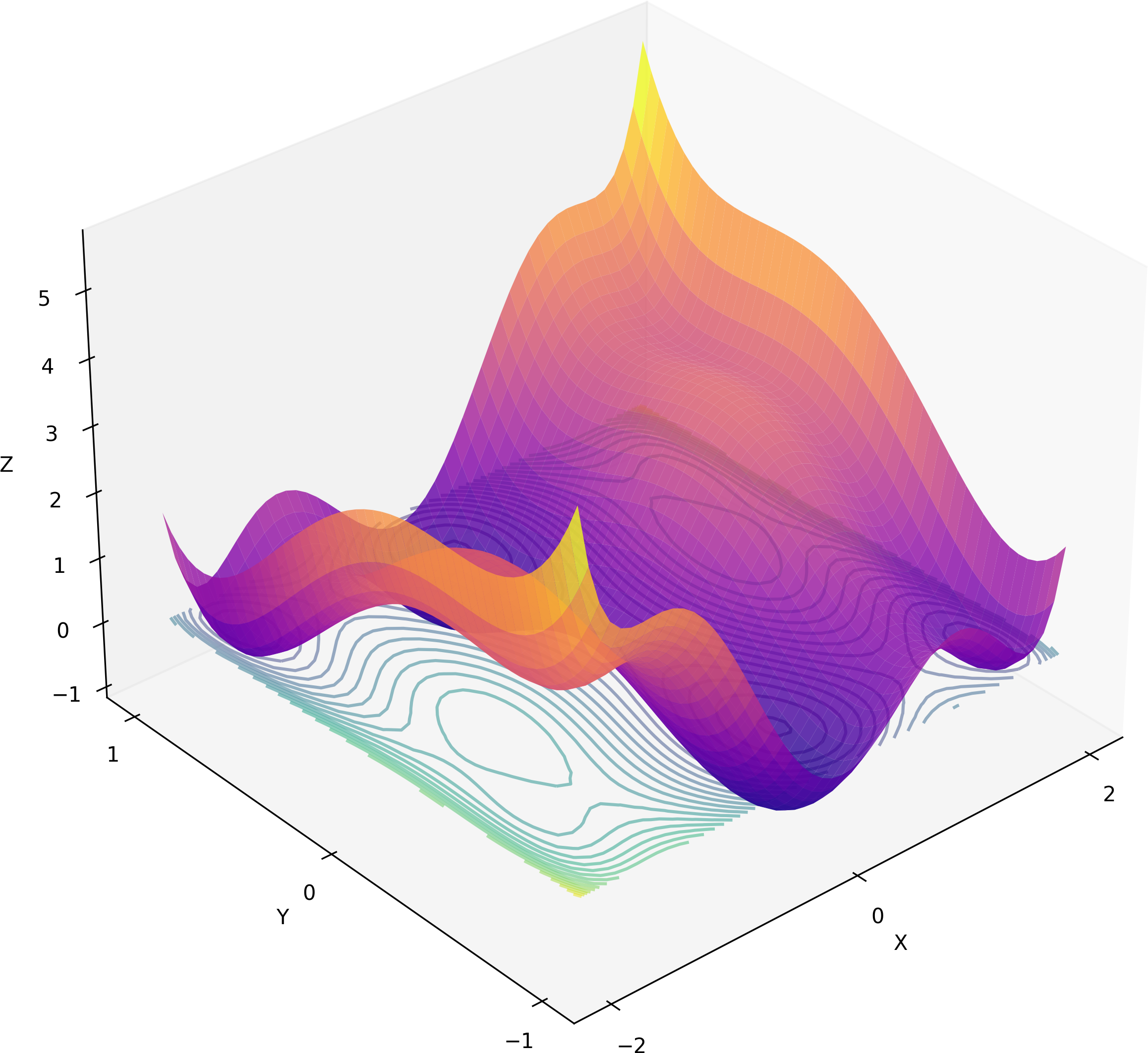} \\[-1mm]
\textbf{Himmelblau} & \textbf{Three Hump Camel} & \textbf{Six Hump Camel}\\[-3mm]
\end{tabular}
\caption{\scriptsize Himmelblau: $f(x,y) = (x^2 + y - 11)^2 + (x + y^2 - 7)^2$. Three-Hump Camel: $f(x,y) = 2 \cdot x^2 - 1.04 \cdot x^4 + \frac{x^6}{6} + xy + y^2$. Six-Hump Camel: $f(x,y) = (4 - 2.1x^2 + \frac{x^4}{3}) \cdot x^2 + xy + (-4 4\cdot y^2) \cdot y^2$.}
\label{fig:SyntheticLandscapes}
\end{figure*}

\begin{figure}

\begin{mdframed}
\vspace*{-1mm}
\noindent\begin{minipage}{\linewidth}
\centering
 \begin{minipage}{.5\linewidth}
 \tiny
 \captionof*{figure}{\small Himelblau}
 \label{funct:Himelblau}
\includegraphics[width=\textwidth]{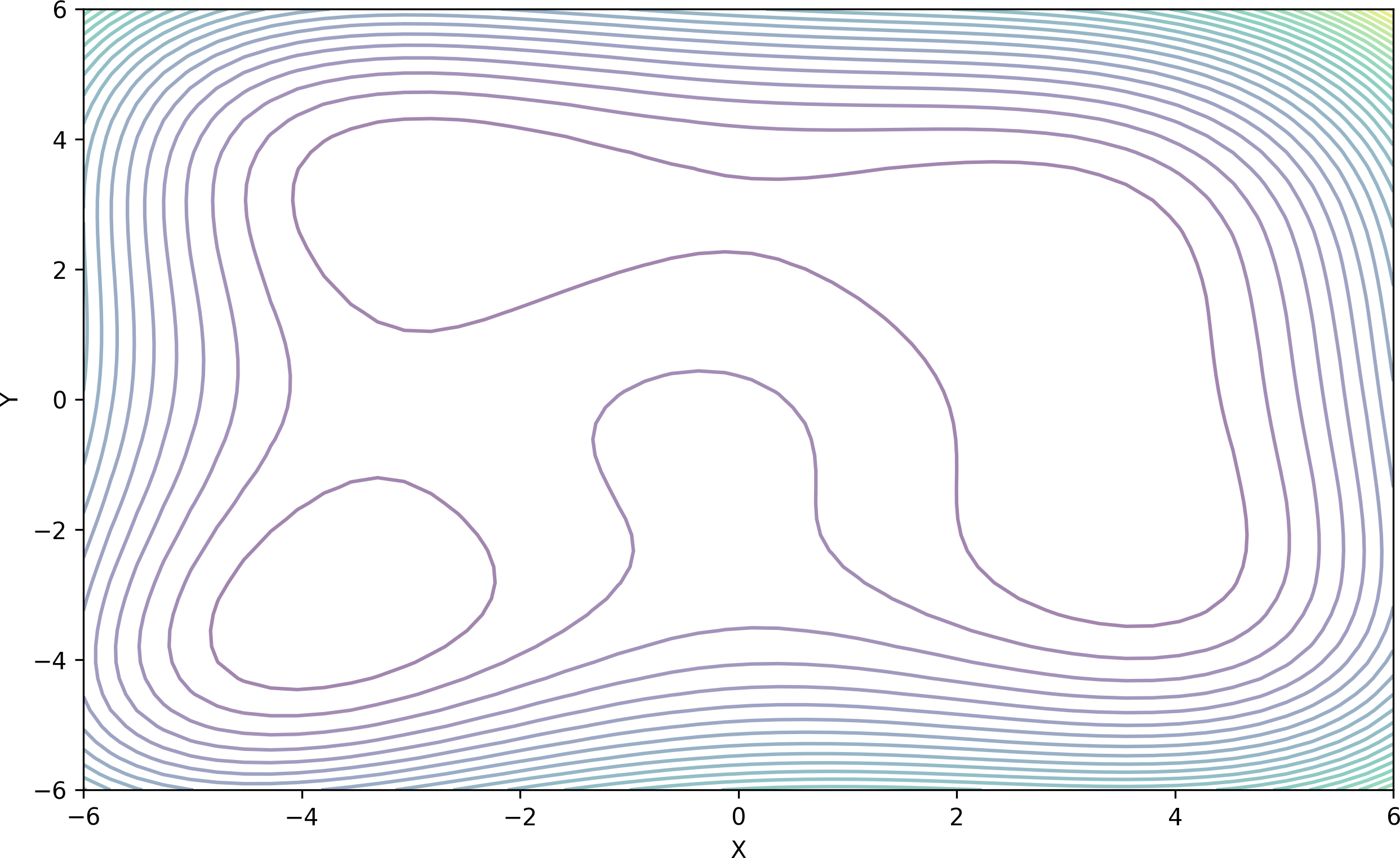}
 \end{minipage}
\hspace*{2mm}
\begin{minipage}{.45\linewidth}
 \centering
 \tiny
The Himmelblau function has four global minima:
\begin{enumerate}
    \item $f(3.0, 2.0) = 0.0$
    \item $f(-2.805118, 3.131312) = 0.0$
    \item $f(-3.779310, -3.283186) = 0.0$
    \item $f(3.584428, -1.848126) = 0.0$
\end{enumerate}
 \end{minipage}
 \hspace*{2mm}
\end{minipage}
\end{mdframed}

\begin{mdframed}
\vspace*{-1mm}
\noindent\begin{minipage}{\linewidth}
\centering
 \begin{minipage}{.5\linewidth}
 \captionof*{figure}{\small Three-Hump Camel}
 \label{funct:ThreeHumpCamel}
\includegraphics[width=\textwidth]{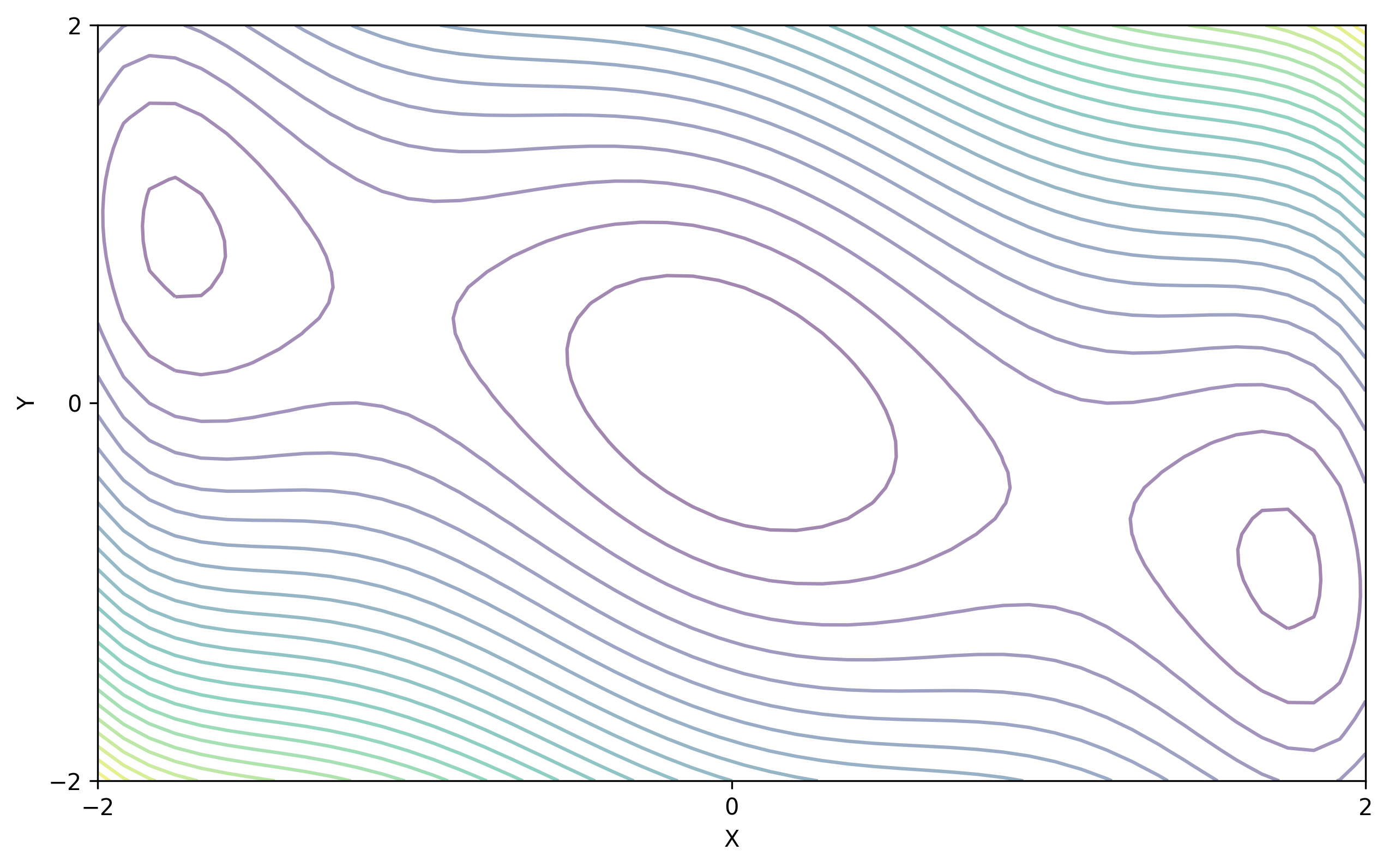}
 \end{minipage}
\hspace*{2mm}
\begin{minipage}{.45\linewidth}
 \centering
  \tiny
The Three-Hump camel function has a global minimum  and two local minima:
\begin{enumerate}
    \item $f(0,0) = 0$
    \item $f(1.7475,-0.8737) \approx 0.2986$    
    \item $f(-1.7475,0.8737) \approx 0.2986$
\end{enumerate}
 \end{minipage}
 \hspace*{2mm}
\end{minipage}
\end{mdframed}

\begin{mdframed}
\vspace*{-1mm}
\noindent\begin{minipage}{\linewidth}
\centering
 \begin{minipage}{.5\linewidth}
\tiny
 \captionof*{figure}{\small Six-Hump Camel}
 \label{funct:SixHumpCamel}
\includegraphics[width=\textwidth]{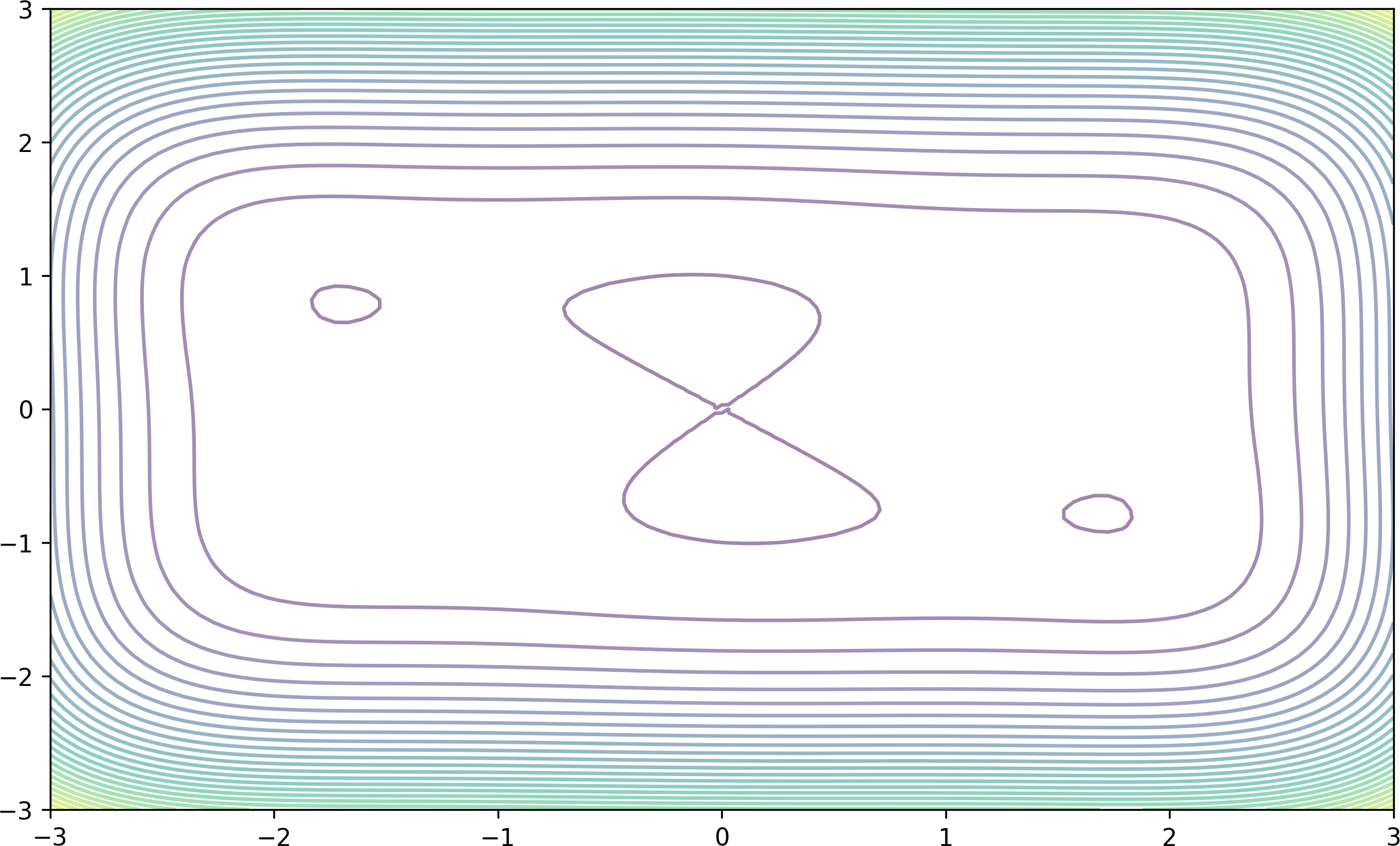}
 \end{minipage}
\hspace*{2mm}
\begin{minipage}{.45\linewidth}
 \centering
  \tiny
The Six-Hump Camel function has two global minima  
 and four local minima:
\begin{enumerate}
    \item $f(-0.0898,0.7126) = -1.0316$
    \item $f(0.0898,-0.7126) = -1.0316$    
    \item $f(-2.8051, -0.0312) \approx 63.848$
    \item $f(0.9805, 1.8367)\approx -11.5 $  
    \item $f(1.8839, -1.5252) \approx -3.14 $ 
    \item $f(-1.8658, 1.4900)\approx -2.64 $  
\end{enumerate}

 \end{minipage}
 \hspace*{2mm}
\end{minipage}
\end{mdframed}

\caption{\scriptsize Locations of global and local minima of Himmelblau, Three-Hump Camel and Six-Hump Camel function}
\label{fig:SyntheticLocalGlobal}

\end{figure}

We first select three complex synthetic functions that capture key characteristics of real-world loss landscapes. They are rich in local minima of varying sharpness, as illustrated in Figure~\ref{fig:SyntheticLandscapes}. Their contour plots are shown in Figure~\ref{fig:SyntheticLocalGlobal}, which also lists present global and local minima.

\subsection*{Optimization Dynamics in a Controlled Setting}

To simplify the exploration of how an optimizer performs, it's beneficial to examine its behavior in a well-defined environment. For this purpose, synthetic functions that possess identifiable global and local minima provide an ideal setting. To achieve this, rather than characterizing a single model, we sample the stationary distribution of an optimizer by ``restarting" it $R$ times, effectively initializing $\w_0$ non-uniformly at random; these instances are also known as random restarts in optimization literature. For each initial condition and for each optimizer, the resulting trajectory (comprising consecutive models, which we typically visualize during training via training loss) continues for a fixed budget of $I$ iterations. The end model of each trajectory is then added to a population of models. The population for each optimizer (consisting of $k$ low-loss, converged models) represents the stationary distribution of that optimizer. This population is analyzed for its coverage of the various known global and local minima of a synthetic landscape, providing us with a comprehensive view of any inherent biases of an optimizer towards particular local minima.

\subsection*{Statistical Characterization on Real-World Problems}

We also characterize optimizers on the following real-world tasks: Cifar 10 and Cifar 100~\cite {krizhevsky2009learning} image classification problem using ResNet-50~\cite {he2016deep}, emotion classification on goEmotions~\cite{demszky2020goemotions} and TweetEval~\cite{barbieri2020tweeteval} datasets. We select these tasks to account for both settings of accuracy or macro-F1 as indicators of the generalizability of an algorithm. We note that in text mining and NLP tasks, macro-F1 is more popular due to data imbalance challenges in multi-class classification tasks. 

In real-world tasks, we lack knowledge of the loss landscape, preventing us from employing the approach described above for synthetic landscapes. Instead, to still accommodate the stochastic nature of an optimizer, we propose the following approach. We sample from a given number $Tr$ of random restarts; here, $Tr < R$ due to the typically higher cost of optimizing on real-world loss landscapes compared to synthetic ones. The key insight lies in treating each trajectory as a local view provided by an optimizer of a loss landscape. Therefore, we sample not just the last/converged model from a trajectory, but $L$ models.

We explore two settings to obtain two distinct populations of models for an optimizer: (1) sampling the $L$ lowest-loss models from a given trajectory; (2) sampling the $L$ models with the highest generalization capability (accuracy or macro-F1 depending on the task). These two populations are then compared via statistical tests to determine whether there are any statistically significant differences, thus providing a global/population view on whether models selected by loss are conducive to high generalization capability. By comparing populations of models, we can also fairly evaluate two given optimizers and avoid drawing conclusions based on arbitrary or hand-selected models. For instance, we can use statistical tests to compare the $L$ lowest-loss models over $Tr$ trajectories of Algorithm A with the $L$ lowest-loss models over $Tr$ trajectories of Algorithm B, testing for statistically significant differences between the populations. Our proxy for ``better" is test set accuracy or macro-F1.

We note that we compare a total of $6$ algorithms: SGD, two noise-enabled variants that have appeared in literature, to which we refer as NiG-GD/SGD and NiG-GD/SGD, SAM, and several BH-based variants described in greater detail later in the paper.

%% file: sections/Basin-Hopping-Algorithms.tex
\section{Broadening Stochastic Optimizers under BH Umbrella: New Noise-Enabled Optimizers}
\label{sec:Algorithms}

The four core algorithms analyzed in this paper are SGD, NoiseInModel-GD/SGD (abbreviated as NiM-GD/SGD), NoiseInGradient-GD/SGD (abbreviated as NiG-GD/SGD), and SAM, as described in Section~\ref{sec:RelatedWork}. The pseudo-codes of the latter are presented in the Related Work section. By enabling noise as NiG or NiM, and varying over BH, MonotonicBH, we obtain four more algorithms, referred to as NiG-BH, NiM-BH, NiG-MBH (`M' for Monotonic), NiM-MBH. These algorithms are instantiations of the BH framework for deep learning optimization.

While not presented in this manner, noise-enabled optimizers combine two components, one that exploits the landscape via a local search (the gradient-based model update) and one that explores the landscape via injecting noise in the gradient or the model. These two are core components of the BH framework, which we respectively name {\tt LocalSearch} and {\tt Perturb}. NoiseInModel or NoiseInGradient  then become specific instantiations of a BH framework, which has a rich history~\cite{OlsonShehuAdvAI12} and has been adapted for multi-basin landscapes of actuated physical and biological systems~\cite{MolloyShehuRobotica16, MaximovaPlakuShehuBIBM15, MaximovaShehuJCB17, maximovaplakushehutcbb17}. 

The framework is related in Algorithm~\ref{alg:BH}; as presented, it permits a more general stopping criterion than a fixed budget $T$. BH iterates between minima $Y_i$ in the parameter space, to which {\tt LocalSearch} maps a point $X_i$. $X_{i > 0}$ are obtained by perturbing the current minimum $Y_i$, and the {\tt Perturbation} component broadens the injection of noise. The distinction between $X$ and $Y$ to present minima clearly, but both correspond to model parameters.

\begin{figure}[htbp]
\vspace*{-3mm}
\begin{mdframed}
\vspace*{-2mm}
\captionof{algorithm}{(Monotonic) BH Framework}
\label{alg:BH}
\vspace*{-2mm}
\begin{algorithmic}[1]
\STATE {\bf Input:} $f(\w)$  
\STATE {\bf Output:} $\w$ 

\STATE $i \gets 0$
\STATE $X_i \gets$ random initial point
\STATE $Y_i \gets LocalSearch(X_i)$
\WHILE{NOT STOP}
    \STATE ${X_{i+1}} \gets Perturb(Y_i)$
    \STATE $Y_{i+1} \gets LocalSearch(X_{i+1})$
    \IF{$f(Y_{i+1}) < f(Y_i)$}
        \STATE $i \gets i+1$
    \ENDIF
    \ENDWHILE
\end{algorithmic}
\end{mdframed}
\vspace*{-2mm}
\begin{mdframed}
\vspace*{-2mm}
\captionof{algorithm}{LocalSearch}
\label{alg:LocalSearch}
\vspace*{-2mm}
\begin{algorithmic}[1]
\STATE {\bf Input:} $f(\w), \w, \tau > 0, \eta, \epsilon$  
\STATE {\bf Output:} $\w, t$
\WHILE{$t < \tau$}
  \STATE $\g \gets \nabla f(\w)$
  \IF{$|\g | < \epsilon$}
      \STATE Terminate
  \ENDIF
  \STATE $\w \gets \w - \eta \cdot \g$
\ENDWHILE
\end{algorithmic}
\vspace*{-1mm}
\end{mdframed}
\vspace*{-2mm}
\begin{mdframed}
\vspace*{-1mm}
\noindent\begin{minipage}{\textwidth}
\centering
\hspace*{-10mm}
 \begin{minipage}{.4\textwidth}
 \centering
 \captionof{algorithm}{PerturbModel}
\label{alg:PerturbModel}
\vspace*{-2mm}
\begin{algorithmic}[1]
\STATE {\bf Input:} $\w, \rho $  
\STATE {\bf Output:} $\w$ 

\STATE ${\bf \zeta} \in B_0(\rho)$ 
\STATE ${\w} \leftarrow \w + {\bf \zeta}$ 
\end{algorithmic}
\end{minipage}
\hspace*{2mm}
\begin{minipage}{.4\textwidth}
 \centering
\captionof{algorithm}{PerturbGradient}
\label{alg:PerturbGradient}
\vspace*{-2mm}
\begin{algorithmic}[1]
\STATE {\bf Input:} $\g, \rho $  
\STATE {\bf Output:} $\g$ 

\STATE ${\bf \zeta} \in B_0(\rho)$ 
\STATE ${\bf g} \leftarrow {\bf g} + {\bf \zeta}$ 

\end{algorithmic}
\end{minipage}

\end{minipage}
\end{mdframed}
\vspace*{-2mm}
\begin{mdframed}
\vspace*{-2mm}
\captionof{algorithm}{NiG-BH}
\label{alg:NoiseInGradient-MonotonicBH}
\vspace*{-2mm}
\begin{algorithmic}[1]
\STATE {\bf Input:} $f(\w), T >0, \epsilon \cong 0, \tau >0, \eta, \rho $ 
\STATE {\bf Output:} $\w$ 

\STATE $(\w,tls) \gets LocalSearch(f, \w, \tau, \eta, \epsilon)$ 
\STATE $t \gets t + tls$

\WHILE{$t \leq T$} 
   \STATE $\g \leftarrow \nabla f(\w_t)$
   \STATE $\g \gets PerturbGradient(\g, \rho)$
   \STATE $\w \gets \w - \eta \cdot g$
   \STATE ($\w, tls) \gets LocalSearch(f, \w, \tau, \eta, \epsilon)$
   \STATE $t \gets t + tls$
\ENDWHILE
\end{algorithmic}
\vspace*{-1mm}
\end{mdframed}
\vspace*{-2mm}
\begin{mdframed}
\vspace*{-2mm}
\captionof{algorithm}{NiM-MBH}
\label{alg:NoiseInModel-MonotonicBH}
\vspace*{-2mm}
\begin{algorithmic}[1]
\STATE {\bf Input:} $f(\w), T >0, \epsilon \cong 0, \tau >0, \eta, \rho $  
\STATE {\bf Output:} $\w$ 

\STATE $(\w,tls) \gets LocalSearch(f, \w, \tau, \eta, \epsilon)$ 
\STATE $t \gets t + tls$

\WHILE{$t \leq T$} 
   \STATE $\w \gets PerturbModel(\w, \rho)$
   \STATE $(\w_{cand}, tls) \gets LocalSearch(f, \w, \tau, \eta, \epsilon)$
   \STATE $t \gets t + tls$
   \IF{$f(\w_{cand}) < f(\w)$}
     \STATE $\w \gets \w_{cand}$     
   \ENDIF
\ENDWHILE
\end{algorithmic}
\vspace*{-1mm}
\end{mdframed}

\vspace*{-4mm}
\end{figure}

Line 9 in Algorithm~\ref{alg:BH} makes this particular presentation monotonic-BH. Removing the condition in line $9$ provides us with the general BH formulation. The BH framework can be easily instantiated for deep learning optimization. The {\tt LocalSearch} is the gradient-based model update (the discrete step in the direction of steepest descent). The {\tt Perturbation} component can be implemented in two different ways, either to inject noise in the gradient or the model directly, resulting in two different instantiations, to which we refer as NiG-BH and NiM-BH, respectively. Note that if monotonicity is enforced (as in line $9$ in Algorithm~\ref{alg:BH}, then one obtains NiG-MBH (shown in Algorithm~\ref{alg:NoiseInGradient-MonotonicBH}) and NiM-MBH (shown in Algorithm~\ref{alg:NoiseInModel-MonotonicBH}). We note that in our implementation, as shown in Algorithm~\ref{alg:LocalSearch}, {\tt LocalSearch} carries out $\tau < T$ iterations of gradient descent, or terminates earlier if the gradient flattens. {\tt PerturbModel}, shown in Algorithm~\ref{alg:PerturbModel} is an implementation of {\tt Perturb} by injecting noise (vector $\zeta$) in the model. The returned model parameter is only a candidate (line 7), given the monotonicity constraint (line 8). Equivalently, injecting noise directly in the gradient can be implemented, shown in PerturbGradient in Algorithm~\ref{alg:PerturbGradient}.

The BH framework is rich and permits various algorithmic instantiations to assess the exploration-exploitation balance. In this paper we debut and analyze the BH and monotonic BH (MBH)frameworks. In each of these, we investigate adding noise in the gradient (through {\tt PerturbGradient}) or in the model (through {\tt PerturbModel}).

%% file: sections/SyntheticLoss-Analysis.tex
\section{Stationary Distribution on Synthetic Nonconvex Loss Landscapes}

Table~\ref{table:SyntheticFunctions-StationaryDistribution-Comparison} shows the stationary distribution (end points of $500$ trajectories, each initiated from a random point sampled uniformly at random over the domain of a function) for each of the 6 algorithms in terms of percentages of converged models over the known and local global minima of the synthetic landscapes. These are the ``base" versions of the algorithms, with no hyperparameter tuning. For each synthetic function, the global minima are listed first, followed by the local minima. Flatter minima are listed before sharper ones.

\begin{table*}[h]
\centering
\tiny
\begin{tabular}{|p{5em}||p{2em}|p{2em}|p{2em}|p{2em}|p{2em}||p{2em}|p{2em}|p{2em}|p{2em}||p{2em}|p{2em}|p{2em}|p{2em}|p{2em}||p{2em}||p{2em}|}
\hline
\textbf{Algorithms} & \multicolumn{5}{c|}{\bf Himmelblau} & \multicolumn{4}{c|}{\bf Three-Hump Camel} & \multicolumn{7}{c|}{\bf Six-Hump Camel}\\
\hline
&  \textbf{GM1} & \textbf{GM2} & \textbf{GM3} & \textbf{GM4} & \textbf{Else} &
   \textbf{GM} & \textbf{LM1} & \textbf{LM2} & \textbf{Else} &
   \textbf{GM1} & \textbf{GM1} & \textbf{LM1} & \textbf{LM2} &\textbf{LM3} & \textbf{LM4} & \textbf{Else}\\\hline

GD &  28 &  25 & 23 &  24 & 0   & 32 & 35 & 33 & 0  & 4.25 &  23.5  &  22.75 & 17.25 &  1.5 &  10.25 & 20.5 \\

NiG-GD & 28 & 23 & 23 & 25 & 1  & 31 & 35 & 34 & 0  & 13.5  &	16.4  &	14.6  &  	15.5 &	0.4  &	1.6 & 38\\

NiM-GD & 30 & 27 & 23 & 20 & 0  & 34 & 34 & 32 & 0  & 7.375  &	18.38 	& 19.88  &	16.5 	& 10   &	9.86 &	17 \\

SAM &  30 &  23 & 20 & 20 & 7   & 30 & 27 & 25 & 18 & 11.63 &16. 1 &	19  &	16.13 &	0.75 &	3 & 34  \\

NiG-BH & 27 & 25 & 25 & 22 & 1  & 32 & 35 & 33 & 0  & 6.25 & 21.75 & 21.75 & 17.75 &   2.   &  8.25 & 22\\

NiM-BH &  27 & 25 & 24 & 24 & 0 & 33 & 35 & 31 & 1  & 6.5 & 20.25 & 21.75 & 17.25&   2.25 & 10.25 & 22\\

NiG-MBH & 22 & 27 & 23 & 28 & 0 & 33 & 33 & 34 & 0  & 5.75 &  23.75 & 24.25 & 16.25  & 2 & 11 & 17\\

NiM-MBH & 30 & 23 & 20 & 20 & 7 & 34 & 32 & 34 & 0  & 5.25 & 24.75 & 24.25 & 16.5 &    1.75 & 11.5 & 12\\

\hline
\end{tabular}
\vspace*{-2mm}\caption{\scriptsize The stationary distribution (reported in \% for each entry) for the Himmelblau, Three-Hump Camel, and Six-Hump Camel function for each algorithm.}
\label{table:SyntheticFunctions-StationaryDistribution-Comparison}
\end{table*}

\begin{figure}[htbp]
\vspace*{-2mm}
\centering
\begin{tabular}{c}
\textbf{Himmelblau}\\
\includegraphics[width=0.49\textwidth]{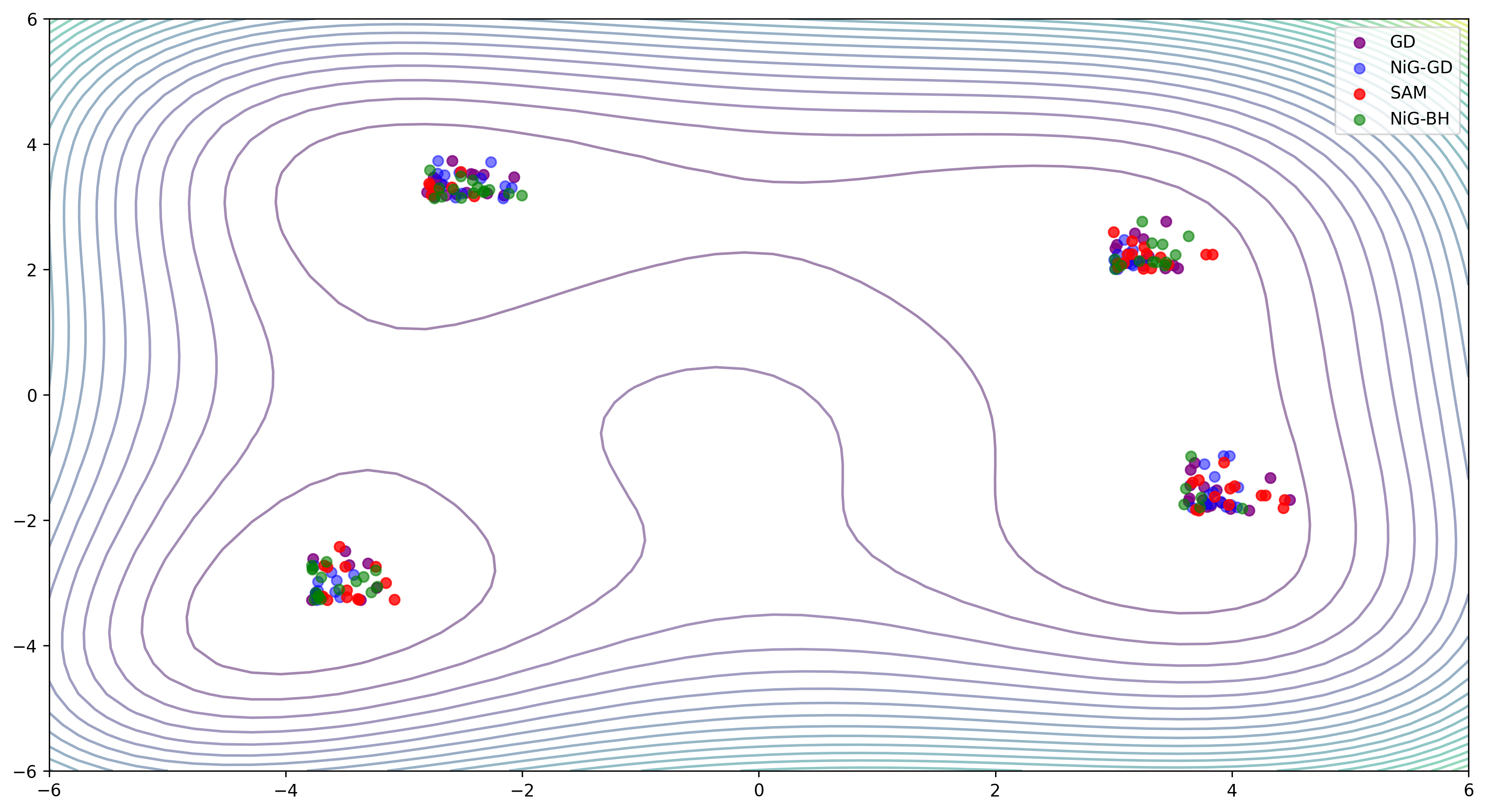}\\
\textbf{Three Hump Camel}\\[2mm]
\includegraphics[width=0.49\textwidth]{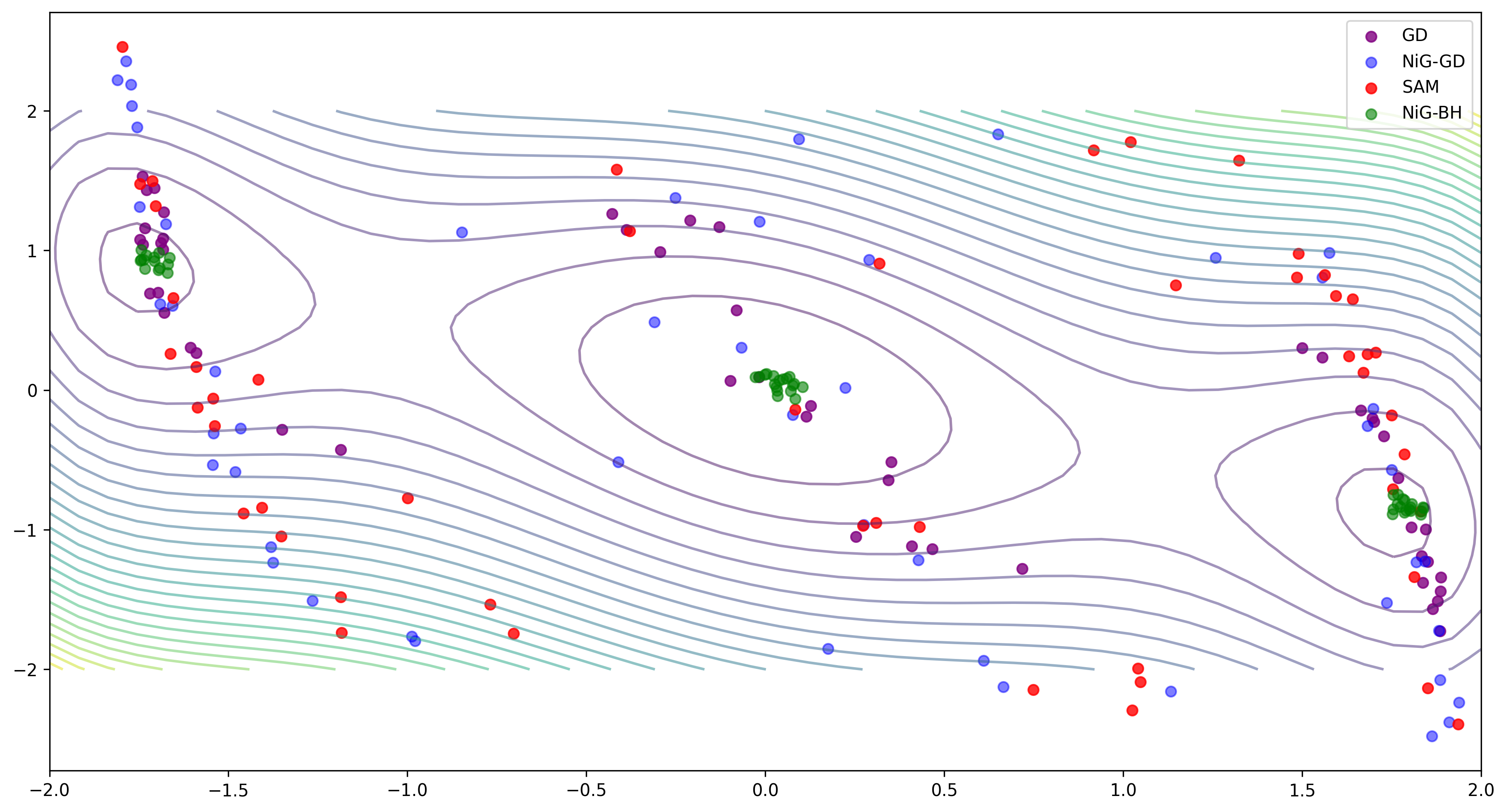} \\
\textbf{Six Hump Camel}\\[2mm]
\includegraphics[width=0.49\textwidth]{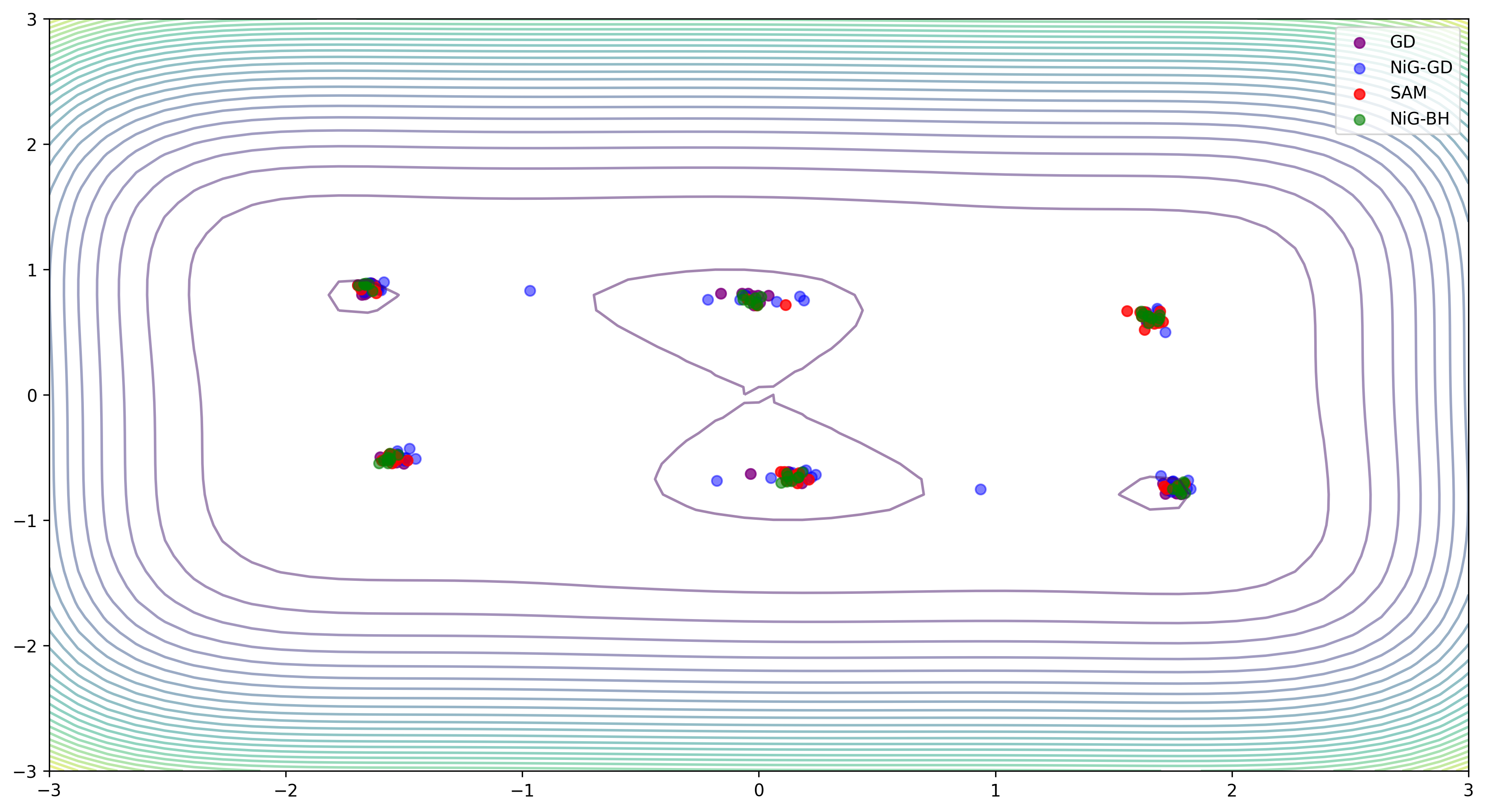} \\
\end{tabular}
\caption{\scriptsize Stationary distribution of the optimization trajectory end-points by GD, NiG-GD~\cite{JinJordan17}, SAM ~\cite{ForetNeyshabur21}, and NiG-BH. Distribution is shown for only 50 trajectories for each algorithm for a clear visual presentation.}
\label{fig:SyntheticLandscapes-SelectedAlgorithms-Visual}
\end{figure}

Figure~\ref{fig:SyntheticLandscapes-SelectedAlgorithms-Visual} shows $50$ end-points (sampled from $500$) of selected algorithms on the three synthetic functions.

Several interesting observations emerge. First, the Six-Hump Camel function presents more challenges for all functions. The population of the first global minimum is low, and the percentage of "non-converged" trajectories is higher (the number of end-points that do not fall in any of the known minima (indicated by 'Else' in the tables). NiG-GD and SAM do particularly poorly on this function, with $38$\% and $34$\% of the stationary distribution respectively not falling in any of the known minima. Interestingly, however, the stationary distribution of these two optimizers is skewed away from LM3 and LM4, which are the sharpest local minima (note that minima are ordered by sharpness, from low to high) in our exposition). Without any particular considerations, just noise in the gradient, NiG-GD matches SAM's performance in skewing the stationary distribution towards the global and flatter minima. This skewness is not observed in the other optimizers, as expected. It is interesting that the BH optimizers have more end-points converging over minima than other regions of the landscape. In the rest of the analysis, we exclude algorithms based on monotonic BH. Their greedy nature, while exploiting well certain synthetic landscapes, makes them prone to premature convergence on complex, real-world landscapes (data not shown), a behavior that is known of this framework in complex optimization problems~\cite{OlsonShehuAdvAI12}.

%% file: sections/Real-World-Analysis.tex
\section{Model Population Analysis on Tasks}

As related earlier, we obtain a population of models that are ``samples" of the view obtained by a particular optimizer of a loss landscape. We set $Tr = 5$ and $L = 10$, so we obtain $50$ models from an optimizer. The computational budget for each optimizer (for one trajectory) is $300$ epochs.

\subsection{Generalization versus Optimization}

Our first analysis compares two sets of populations, SetA and SetB. SetA is the population of $50$ lowest-loss models; for each trajectory, the $10$ lowest-loss models are selected. In SetB, from each trajectory, the $10$ highest accuracy (or highest macro-F1) models are selected. These two sets are first compared in terms of accuracy (or macro-F1 depending on the task) via the the Mann-Whitney U test (we utilize {\tt scipy.stats.ttest\_ind} and {\tt scipy.stats.mannwhitneyu} and the two-sided t-test. Both test the null hypothesis that that the distribution underlying SetA is the same as the distribution underlying SetB. While the t-test is parametric, the Mann-Whitney U test is nonparametric. We note that the Mann-Whitney U test is a good choice when the data is not normally distributed and there are no ties in the data (as opposed to the Wilcoxon test which makes both assumptions). 

Table~\ref{table:RealWorld-Mann-Whitney-U-Test-TunedAlgorithms} reports results of the Mann-Whitney U test for the hyperparameter-tuned algorithms, whereas Table~\ref{table:RealWorld-Mann-Whitney-U-Test-BaseAlgorithms} does so for the base algorithms. Table~\ref{table:RealWorld-Mann-Whitney-U-Test-TunedAlgorithms} shows that with few exceptions (NiM-SGD and NiG-BH on GoEmotions-BERT task), the null hypothesis cannot be rejected. That is, one cannot reject that the distribution underlying SetA (models selected by loss) is the same as the distribution underlying SetB (models selected by test set accuracy/macro-F1). The results on the base versions of the algorithms in Table~\ref{table:RealWorld-Mann-Whitney-U-Test-BaseAlgorithms} support this finding. The t-test, related in Table~\ref{table:RealWorld-T-Test-TestSetPerformance} provides additional support.

\begin{table}[htbp]
\scriptsize
  \centering
  \begin{subtable}{0.4\textwidth}
    \centering
\begin{tabular}{|p{5em}||p{3.6em}|p{3.6em}|p{3.3em}|p{3.9em}|}
\hline
\textbf{Algorithm} & Cifar10 Resnet50 & Cifar100 Resnet50 & GoEmot & TweetEval\\
\hline
SGD & 0.0821 & 0.1941 & 0.4192 & 0.1359\\
NiG-SGD & 0.4231 & 0.2519 & 0.3618 & 0.4532\\
NiM-SGD & 0.17432 & 0.34121 & {\bf 0.03489} & 0.1837\\
SAM & 0.0915 & 0.051783	& 0.2638 & 0.1834\\
NiG-BH & 0.07532 & 0.6739 & {\bf 0.04868} & 0.4839\\
NiM-BH & 0.18346 & 0.29734 & 0.18942 & 0.3574\\
\hline
\end{tabular}
    \caption{\scriptsize P-values are reported for the Mann-Whitney U test when comparing SetA to SetB on test set performance for each hyperparameter-optimized algorithm over each of the real-world tasks.}
    \label{table:RealWorld-Mann-Whitney-U-Test-TunedAlgorithms}
  \end{subtable}
  \hfill
  \begin{subtable}{0.4\textwidth}
    \centering
\begin{tabular}{|p{5em}||p{3.6em}|p{3.6em}|p{3.3em}|p{3.9em}|}
\hline
\textbf{Algorithm} & Cifar10 Resnet50 & Cifar100 Resnet50 & GoEmot & TweetEval\\
\hline
SGD & 0.2315 & 0.19332 & 0.2875 & 0.4621\\
NiG-SGD & 0.2989 & 0.5429 & 0.4632 & 0.1654\\
NiM-SGD	& 0.6543 & 0.7563 & 0.6129 & 0.3219\\
SAM	& 0.0978 & {\bf 0.01073} & 0.1984	& 0.2861\\
NiG-BH & 0.3569	& {\bf 0.0285} & 0.8328 & 0.3951\\
NiM-BH & 0.6153	& {\bf 0.0472} & 0.6143 & 0.5178\\
\hline
\end{tabular}
    \caption{\scriptsize P-values for the Mann-Whitney U test comparing SetA to SetB on test set performance for each base algorithm over each real-world task. }
\label{table:RealWorld-Mann-Whitney-U-Test-BaseAlgorithms}
  \end{subtable}
  \caption{\scriptsize Results on the Mann Whitney U Test. P-values $< 0.05$ are highlighted in bold font.}
  \label{tab:RealWorld-Mann-Whitney-U-Test}
\end{table}


\begin{table}[htbp]
\scriptsize
  \centering
  \begin{subtable}{0.4\textwidth}
    \centering
    \begin{tabular}{|p{5em}||p{3.6em}|p{3.6em}|p{3.3em}|p{3.9em}|}
\hline
\textbf{Algorithm} & Cifar10 Resnet50 & Cifar100 Resnet50 & GoEmot & TweetEval\\
\hline

SGD	& 0.2314 & 0.3156 & 0.7563 & 0.54623\\
NiG-SGD	& 0.4961 & 0.5753 & 0.72134	& 0.1823\\
NiM-SGD	& {\bf 0.0421} & 0.1291  & 0.2961 & 0.1962\\
SAM	& 0.7532 & 0.6982 & 0.6432 & 0.5391\\
NiG-BH & 0.3612	& 0.18326 & {\bf 0.03135} & 0.1837\\
NiM-BH & 0.6318	& 0.1938 & 0.7128 & 0.8723\\
\hline
\end{tabular}
    \caption{\scriptsize P-values for the t-test when comparing SetA to SetB for each hyperparameter-optimized algorithm over each of the real-world tasks. }
    \label{table:RealWorld-T-Test-TunedAlgorithms}
  \end{subtable}
  \hfill
  \begin{subtable}{0.4\textwidth}
    \centering
    \begin{tabular}{|p{5em}||p{3.6em}|p{3.6em}|p{3.3em}|p{3.9em}|}
\hline
\textbf{Algorithm} & Cifar10 Resnet50 & Cifar100 Resnet50 & GoEmot & TweetEval\\
\hline
SGD	& 0.8764 & 0.14019 & 0.2345 & 0.5972\\
NiG-SGD	& 0.8195 & 0.8423 & 0.8744 & 0.4426\\
NiM-SGD	& 0.8345 & 0.7425 & 0.34556 & 0.4585\\
SAM	& 0.7213 & 0.76894 & 0.6754 & 0.6764\\
NiG-BH & 0.5678 & {\bf 0.01245} & 0.45354 & 0.53254\\
NiM-BH & 0.9134 & 0.8325 & 0.3958 & 0.6467\\
\hline
\end{tabular}
    \caption{\scriptsize P-values are reported for the t-test when comparing SetA to SetB for each base algorithm over each of the real-world tasks.}
    \label{table:RealWorld-T-Test-BaseAlgorithms}
  \end{subtable}
  \caption{\scriptsize Results on the t-test. P-values less than $0.05$ are highlighted in bold font.}
  \label{table:RealWorld-T-Test-TestSetPerformance}
\end{table}


In Table~\ref{table:RealWorld-Accuracies-F1-TunedAlgorithms}, we present the average and standard deviation of the test error accuracy or macro-F1 for SetA (low-loss models, top row for each optimizer) versus SetB (high generalization models, bottom row for each optimizer) for each optimizer. Here, we narrow our focus to the hyperparameter-optimized optimizers. Comparison across optimizers over SetA and SetB reveals comparable accuracies and standard deviations. However, some interesting observations emerge. When focusing on SetA, the set of low-loss models, and comparing the average (test) accuracies/macro-F1s for each algorithm, we observe that on the accuracy-evaluated tasks, such as CIFAR 10 and CIFAR 100, the top three optimizers (with the highest three accuracies) are SGD (twice), NiM-SGD (once), and  SAM (twice). On the macro-F1-evaluated tasks, such as GoEmotions (BERT) and TweetEval (BERT), the top three optimizers (with the highest three macro-F1s) are SGD (once), NiM-SGD (once), and NiM-BH (twice). SAM slightly underperforms on the macro-F1 tasks, while the BH-based optimizers seem to have a slight advantage.

\begin{table}[htbp]
\centering
\tiny
\begin{tabular}{|p{5em}||p{6em}|p{6em}|p{6em}|p{6em}|}
\hline
\textbf{Algorithm} & Cifar10 Resnet50 & Cifar100 Resnet50 & GoEmot & TweetEval\\
\hline

\multirow{2}{*}{SGD} & ({\bf 0.934},0.004)   & ({\bf 0.776}, 0.021) & (0.493, 0.032) & ({\bf 0.599}, 0.025)\\ 
                     & (0.929,0.002) & (0.785, 0.021)  & (0.501, 0.029) & (0.609, 0.019)\\\hline

\multirow{2}{*}{NiG-SGD} & (0.915, 0.004) & (0.759, 0.029) & (0.485, 0.051) & (0.572 , 0.037)\\
                         & (0.918, 0.004) & (0.763, 0.018) & (0.482, 0.049) & (0.579 , 0.032)\\\hline

\multirow{2}{*}{NiM-SGD} & (0.917, 0.005) & ({\bf 0.779}, 0.027) & ({\bf 0.501}, 0.044)  & (0.594, 0.029)\\
                         & (0.925, 0.004) & (0.786, 0.018) & (0.509, 0.039) & (0.596, 0.028)\\\hline

\multirow{2}{*}{SAM} & ({\bf 0.924}, 0.017) & ({\bf 0.779}, 0.037) & (0.459, 0.041) & (0.589, 0.037)\\
                     & (0.941, 0.007) & (0.793, 0.015) & (0.482, 0.023) & (0.595, 0.017)\\\hline

\multirow{2}{*}{NiG-BH} & (0.908, 0.005) & (0.743, 0.019) & (0.486, 0.042) & (0.579, 0.031)\\
                        & (0.912, 0.003) & (0.753, 0.015) & (0.495, 0.036) & ( 0.581, 0.298)\\\hline

\multirow{2}{*}{NiM-BH} & (0.896, 0.019) & (0.749, 0.024) & ({\bf 0.503}, 0.053) & ({\bf 0.602}, 0.035)\\
                        & (0.903, 0.009) & (0.759, 0.022) & (0.506, 0.038) & (0.607, 0.327)\\\hline




\hline
\end{tabular}
\vspace*{-1mm}\caption{\scriptsize For each optimizer, we relate the median accuracy and median standard deviation over SetA (top row for each optimizer) and SetB (bottom row for each optimizer). '(, )' relates '(median accuracy, median standard deviation)' over models in a set. On the last two tasks, reported summary statistics are for macro-F1.}
\label{table:RealWorld-Accuracies-F1-TunedAlgorithms}
\end{table}

We now compare SetA to SetB in terms of training loss in Table~\ref{tab:RealWorld-T-Test-TrainingLoss}. We restrict our attention to the hyperparameter-optimized algorithms. Table~\ref{table:RealWorld-Mann-Whitney-U-Test-TunedAlgorithms-Loss} reports the results of the Mann-Whitney U test statistical test, and Table~\ref{table:RealWorld-T-Test-TunedAlgorithms-Loss} reports the results of the t-test.

\begin{table}[htbp]
\scriptsize
  \centering
  \begin{subtable}{0.4\textwidth}
    \centering
\begin{tabular}{|p{5em}||p{3.6em}|p{3.6em}|p{3.3em}|p{3.9em}|}
\hline
\textbf{Algorithm} & Cifar10 Resnet50 & Cifar100 Resnet50 & GoEmot & TweetEval\\
\hline
SGD & 0.2164 & \textbf{0.0021}& 0.2123 & 0.2952\\

NiG-SGD & 0.092 & 0.2385 & 0.0615 & 0.3152\\

NiM-SGD & 0.1574 & 0.0612 & 0.1286 & 0.2032\\

SAM & \textbf{0.0001} & \textbf{0.0048	}& 0.3810 & 0.2357\\

NiG-BH & \textbf{0.02858} & 0.1426 & \textbf{0.0318} & 0.5412\\

NiM-BH & 0.3745 & 0.1854 & \textbf{0.0325} & 0.2548\\


\hline
\end{tabular}
    \caption{\scriptsize P-values are reported for the Mann-Whitney U test when comparing the loss distributions of SetA to SetB for each algorithm over each of the real-world tasks. P-values less than $0.05$ are highlighted in bold font.}
    \label{table:RealWorld-Mann-Whitney-U-Test-TunedAlgorithms-Loss}
  \end{subtable}
  \hfill
  \begin{subtable}{0.4\textwidth}
    \centering
\begin{tabular}{|p{5em}||p{3.6em}|p{3.6em}|p{3.3em}|p{3.9em}|}
\hline
\textbf{Algorithm} & Cifar10 Resnet50 & Cifar100 Resnet50 & GoEmot & TweetEval\\
\hline
SGD & 0.5631 & 0.7415& 0.3534 & 0.1983\\

NiG-SGD & 0.6512 & 0.1853 & {\bf 0.0916} & 0.4325\\

NiM-SGD & 0.3259 & 0.7122 & 0.3214 & {\bf 0.0851}\\

SAM & {\bf 0.0015} & {\bf 0.0384} & 0.1120 & 0.2352\\

NiG-BH & 0.1523 & 0.2854 & 0.3214 & {\bf 0.0254}\\

NiM-BH & 0.2145 & 0.3847 & {\bf 0.0978} & 0.3021\\


\hline
\end{tabular}
    \caption{\scriptsize P-values are reported for the T test when comparing the loss distributions of SetA to SetB for each algorithm over each real-world task. P-values less than $0.05$ are highlighted in bold font.}
    \label{table:RealWorld-T-Test-TunedAlgorithms-Loss}
  \end{subtable}
  \caption{\scriptsize Results on t-tests on the hyperparameter-optimized (left) and base (right) algorithms. }
  \label{tab:RealWorld-T-Test-TrainingLoss}
\end{table}

\subsection{Population-based Comparison of Optimizers}

We compare pairs of optimizers. Instead of picking one model, we compare the SetA, the population of low-loss models sampled by an algorithm/optimizer over the loss landscape of a real-world task to the population of low-loss models sampled by another algorithm over the same landscape. They are compared here on their test accuracy or macro-F1. To test for differences between the resulting distributions, we utilize the Mann-Whitney U with the null hypothesis that the values (accuracy or macro-F1 depending on the task) in one group (Set A, algorithm X) are the same as the values in the other group (SetA, algorithm Y) Table~\ref{table:PairwiseComparisons} reports the p-values for three pairs of algorithms being compared, SGD vs. SAM, SGD vs. NiM-BH, and SAM vs. NiM-BH. All p-values are higher than $0.05$, which means that the null hypothesis cannot be rejected. This is a profound observation. It suggests that when when expanding our view to a population of low-loss models obtained over several optimization trajectories, we cannot distinguish in performance between SGD and SAM or even between noise-enabled variants of SGD under the BH framework, contradicting observations of SAM superiority made on a hand-selected model.

\begin{table}[htbp]
\centering
\tiny
\begin{tabular}{|p{5em}||p{6em}|p{6em}|p{6em}|p{6em}|}
\hline
\textbf{Task} &	\textbf{SGD vs. SAM} & \textbf{SGD vs. NiM-BH} & \textbf{SAM vs. NiM-BH}\\\hline
Cifar10	      & 0.32458	             & 0.6542	               & 0.0574\\
Cifar100      &	0.47451	             & 0.1247	               & 0.0458\\
GoEmot	      & 0.035478	         & 0.1985	               & 0.1749\\
TweetEval	  & 0.23147	             & 0.3254	               & 0.2158\\
\hline
\end{tabular}
\vspace*{-1mm}\caption{\scriptsize p-values are reported for the Mann Whitney U test of the null hypothesis that two distributions are the same. The distributions are accuracies or macro-F1s (last two tasks) of SetA, low-loss models of an algorithm/optimizers. }
\label{table:PairwiseComparisons}
\end{table}

\paragraph{Learning Curves:}

\begin{figure}[htbp]
\centering
\begin{tabular}{c}
\textbf{Cifar10}\\
\includegraphics[width=0.3\textwidth]{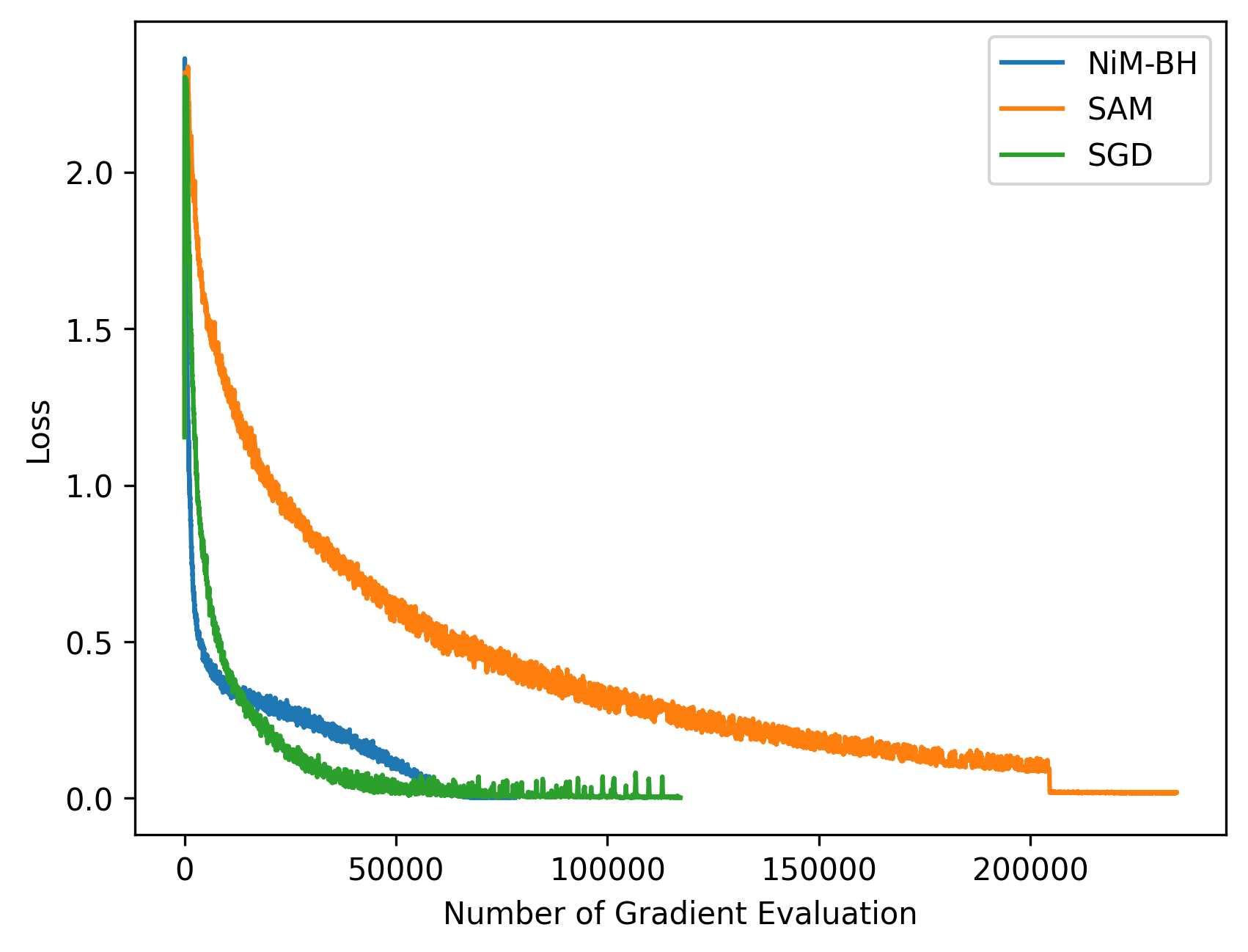}\\[1mm]
\textbf{Cifar100}\\
\includegraphics[width=0.3\textwidth]{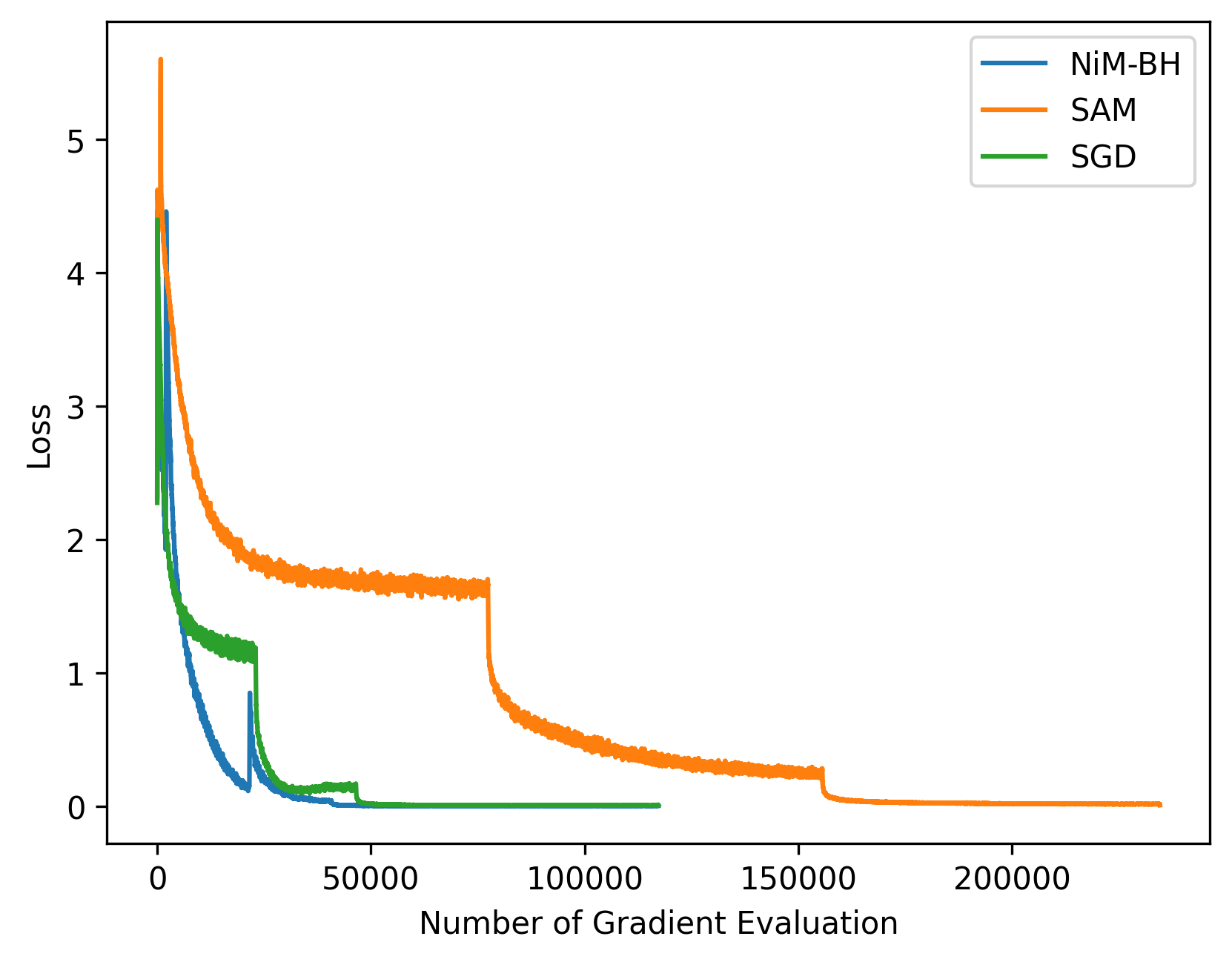}\\[-3mm]
\end{tabular}
\caption{\scriptsize Learning curves for SGD, SAM, and NiM-BH. The y axis shows the smoothed loss (moving average) and the x-axis shows the gradient evaluations.}
\label{fig:LearningCurves}
\end{figure}

Figure~\ref{fig:LearningCurves} shows a learning curve (one representative trajectory) for the three algorithms compared above, SGD, SAM, and NiM-BH. It is evident that SAM spends twice the number of gradient evaluations. If restricted to the same number of gradient evaluations as SGD and NiM-BH, its cannot reach low loss.

%% file: sections/Limitations-Future-Work.tex
\section{Limitations and Future Work}
\label{sec:Limitations-Future-Work}

While some attempt was made to optimize hyperparameters for each algorithm on a real-world task, this greatly increases the computational cost of benchmarking. In future work we plan to grow the profiling of an increasing number of optimizers on more synthetic functions and real-world tasks. More substantially, we believe a fruitful direction of future research includes studying the impact of noise (magnitude of $\rho$ noise vector) in noise-enabled optimizers and its relationship with other hyperparameters for possibly combined effects on the performance of an optimizer. In addition, we believe that noise-enabled optimizers and BH-based algorithms may provide interesting mechanisms to control for low loss, flatness, and other desired characteristics via which researchers can better understand and control for the relationship between better optimization and higher generalization capability.

%% file: sections/Conclusion.tex
\section{Conclusion}
\label{sec:Conclusion}

In this paper we account for the inherent stochastic nature of SGD, noise-enabled variants, and flat-minima optimizers. Broadening noise-enabled SGD variants under the Basin Hopping framework, we present several novel algorithms. We propose a population-based approach into benchmarking these algorithms to better characterize them and more broadly contribute to our understanding of the relationship between optimization and generalization. The primary insight we leverage is that during the training of a neural network, an optimization trajectory grows in a possibly complex nonconvex, multi-dimensional loss landscape and so to characterize for the behavior of an optimizer one needs a nonlocal view that extends over several trajectories and goes beyond the "converged"/lowest-loss model. Synthetic landscapes allow us to rigorously characterize (and compare) the stationary distributions of optimizers. Populations of low-loss models sampled over several optimization trajectories allow us to relate between optimization and generalization and additionally compare optimizers in the statistical sense. Our paper reveals several findings on the relationship between training loss and hold-out accuracy, the comparable performance of SGD, noise-enabled variants, and novel optimizers based on the BH framework; indeed, these algorithms match the performance of flat-minima optimizers such as SAM with half the gradient evaluations. We hope that this work will support further research in deep learning optimization relying not on single models but instead accounting for the stochasticity of optimizers.